\documentclass[letterpaper]{article} % DO NOT CHANGE THIS
\usepackage{aaai25}  % DO NOT CHANGE THIS
\usepackage{times}  % DO NOT CHANGE THIS
\usepackage{helvet}  % DO NOT CHANGE THIS
\usepackage{courier}  % DO NOT CHANGE THIS
\usepackage[hyphens]{url}  % DO NOT CHANGE THIS
\usepackage{graphicx} % DO NOT CHANGE THIS
\usepackage{natbib}  % DO NOT CHANGE THIS AND DO NOT ADD ANY OPTIONS TO IT
\usepackage{caption} % DO NOT CHANGE THIS AND DO NOT ADD ANY OPTIONS TO IT
\usepackage{algorithm}
\usepackage{algorithmic}

\usepackage{xcolor}
\usepackage{graphicx}
\usepackage{amsmath}
\usepackage{amssymb}
\usepackage[bb=dsserif]{mathalpha}
\usepackage{bm}

\usepackage{booktabs,colortbl,tabularx}
\usepackage{pifont}%

\usepackage{mathtools}
\usepackage{commath}

\usepackage{multirow}
\usepackage{comment} 
\usepackage{xspace}

\newlength\savewidth
\newlength\thinwidth

\definecolor{Gray}{gray}{0.93}

\usepackage{xparse}
\usepackage{pifont}
\usepackage{bm}
\usepackage{xspace} % for abbreviation macros
\makeatletter
\DeclareRobustCommand\onedot{\futurelet\@let@token\@onedot}
\def\@onedot{\ifx\@let@token.\else.\null\fi\xspace}
\def\eg{\emph{e.g}\onedot} 
\def\ie{\emph{i.e}\onedot}

\makeatother

\definecolor{Gray}{gray}{0.9}

\usepackage{url}
\usepackage{xcolor}

\newcommand{\red}[1]{{\color{red} #1}}

\newcommand{\cmark}{\ding{51}}%
\newcommand{\xmark}{\ding{55}}%

\usepackage{newfloat}
\usepackage{listings}
\DeclareCaptionStyle{ruled}{labelfont=normalfont,labelsep=colon,strut=off} % DO NOT CHANGE THIS
\floatstyle{ruled}
\newfloat{listing}{tb}{lst}{}
\floatname{listing}{Listing}
\title{The Dynamic Duo of Collaborative Masking and Target for Advanced \\ Masked Autoencoder Learning}
\author{
    Shentong Mo
}
\affiliations{
    Carnegie Mellon University
}

\usepackage{bibentry}
\begin{document}

\maketitle

\begin{abstract}

Masked autoencoders (MAE) have recently succeeded in self-supervised vision representation learning.
Previous work mainly applied custom-designed (\eg, random, block-wise) masking or teacher (\eg, CLIP)-guided masking and targets.
However, they ignore the potential role of the self-training (student) model in giving feedback to the teacher for masking and targets.
In this work, we present to integrate Collaborative Masking and Targets for boosting Masked AutoEncoders, namely CMT-MAE.
Specifically, CMT-MAE leverages a simple collaborative masking mechanism through linear aggregation across attentions from both teacher and student models.
We further propose using the output features from those two models as the collaborative target of the decoder.
Our simple and effective framework pre-trained on ImageNet-1K achieves state-of-the-art linear probing and fine-tuning performance.
In particular, using ViT-base, we improve the fine-tuning results of the vanilla MAE from 83.6\% to 85.7\%.

\end{abstract}

\section{Introduction}

Masked autoencoders (MAE)~\cite{he2021masked} have recently achieved advanced success in learning meaningful visual representations for many downstream tasks, \eg, image classification, object detection, and semantic segmentation.
Meanwhile, researchers also introduced diverse masking pipelines to show the effectiveness of masked modeling in learning meaningful representations from video~\cite{tong2022videomae,MaskedAutoencodersSpatiotemporal2022}, audio~\cite{Huang2022MaskedAT}, and MRI/CT scans~\cite{Chen2023MaskedIM}.

\begin{figure}[!hbt]
\centering
% \fbox{\rule{0pt}{2in}
% \rule{0.8\linewidth}{0pt}}
\includegraphics[width=0.86\linewidth]{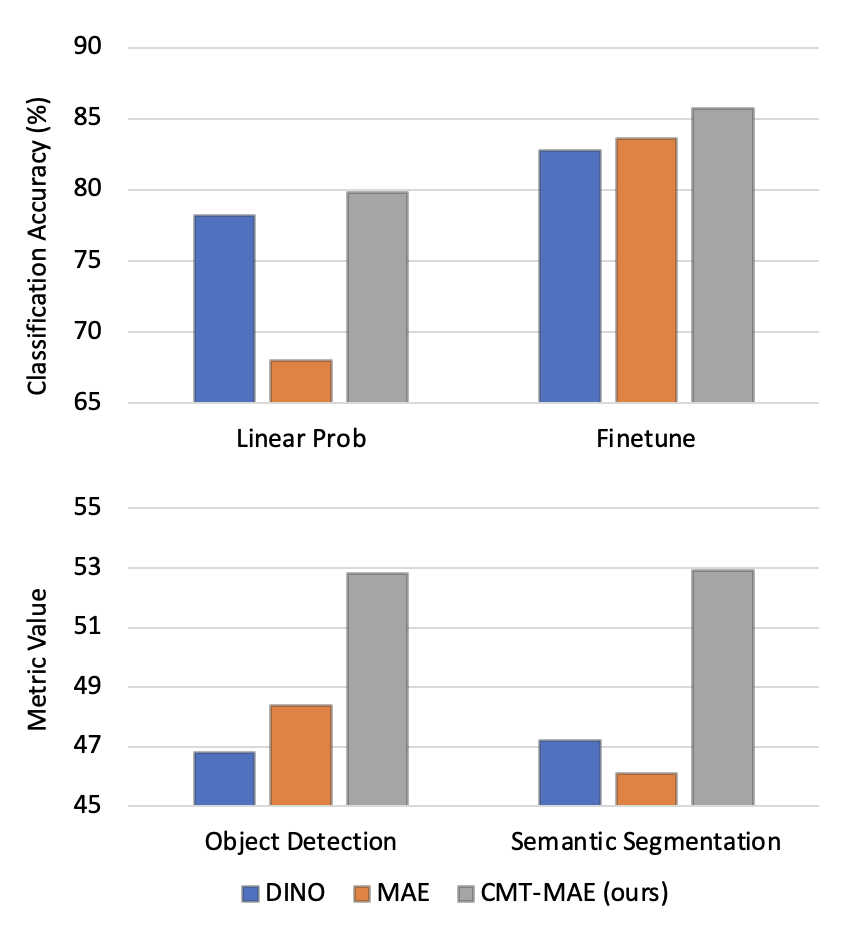}
\caption{{\bf Comparison of our CMT-MAE with MAE and DINO on pre-trained ViT-B/16.} 
Our method significantly outperforms previous baselines in terms of all downstream tasks.}
\label{fig: title_img}
\end{figure}

Exploring masking strategies (\ie, pretext tasks) and supervision targets is critical for MAE~\cite{he2021masked} to capture meaningful features during pre-training.
In this work, we aim to simultaneously improve the powerfulness of the pre-text task and target in MAE-based pre-training on images for self-supervised vision representation learning, which boosts the performance of several downstream tasks compared to MAE~\cite{he2021masked} and DINO~\cite{caron2021emerging}, as shown in Figure~\ref{fig: title_img}.

Early works~\cite{bao2021beit,atito2021sit,he2021masked} on masked image modeling (MIM) mainly applied custom-designed (\ie, random, block-wise, square-wise) masking during pre-training.
For instance, BEiT~\cite{bao2021beit} used a block-wise masking strategy to reconstruct discrete tokens of masked image patches for pre-training transferrable visual representations.
To simplify masked image encoding, the seminal work, MAE~\cite{he2021masked} directly reconstructed missing pixels of 75\% masked patches.
MaskFeat~\cite{wei2022masked} studied five different types of features as supervision targets and observed that Histograms of Oriented Gradients (HOG) with local contrast normalization achieved the best performance.
SimMIM~\cite{xie2022SimMIM} applied a large masked patch size to randomly mask the input image, where RGB values of raw pixels are recovered by a one-layer prediction head.

Recent researchers~\cite{li2021mst,shi2022adversarial,wei2022mvp,li2022semmae} started to leverage a teacher network or adversarial learning to generate the mask and supervision target.
Benefiting from CLIP~\cite{radford2021learning} pre-trained on 400 million image-language pairs, 
MVP~\cite{wei2022mvp} introduced knowledge from CLIP as guidance to achieve impressive gains for MIM-based self-supervised visual pre-training.
AttMask~\cite{kakogeorgiou2022attmask} applied an attention map generated from a teacher transformer encoder, \ie, iBoT~\cite{zhou2022ibot} to guide masking for the student pre-training.
Similarly, SemMAE~\cite{li2022semmae} started to integrate semantic-guided masks from a self-supervised part-learning module with diversity constraints on attention.

While the aforementioned methods achieve promising results, they ignore the potential role of the self-training (student) model in cooperating with the teacher for collaborative masking and targets.
The main challenge is that the teacher and students have different knowledge levels.
The teacher network (\eg, CLIP) masters the knowledge at a higher level than students at initial training, while the student network starts to increase its level across the training. 
To address the aforementioned challenge, our key idea is to simultaneously incorporate two different knowledge-level models (\ie, teacher and student) to guide masking and generate reconstruction targets. 
During training, we aim to leverage students with self-training knowledge to help the teacher with fixed knowledge to guide MIM-based image pre-training dynamically and powerfully.

To this end, we propose a novel masked autoencoder that can integrate the student and teacher networks for collaborative masking and targets, namely CMT-MAE.
In particular, we introduce a simple collaborative masking mechanism through linear aggregation across attention maps from both teacher and student models, which improves the powerfulness of guided masks.
To further boost the performance of downstream tasks, the proposed framework selects representations generated from those two models as the collaborative target for the decoder during pre-training.

Our pre-training process is composed of two stages:
In the first stage, a teacher transformer encoder (\ie, CLIP) takes an input image to extract an attention map from the last attention layer to guide masking.
The student encoder generates features from unmasked patches, which are concatenated with masked tokens to feed into a decoder for recovering the teacher features of masked patches.
In the second stage, we apply a student momentum encoder to generate a student-guided attention map and linearly aggregate it with a teacher-guided attention map to produce the collaborative attention map with a collaborative ratio for collaborative masking.
Then masked tokens concatenate with features of unmasked patches from the student encoder to feed into the decoder.
Finally, two predicted heads are linearly applied to reconstruct the teacher and student features of masked patches for collaborative targets.
It should be noted that the collaborative ratio is also applied to calculate collaborative losses from the teacher and student targets.

Experimental results on ImageNet-1K, MS-COCO, ADE-20K, and DAVIS 2017 demonstrate the state-of-the-art performance of our CMT-MAE. 
In particular, using the backbone of the ViT-base, we improve the fine-tuning results of the vanilla MAE from 83.6\% to 85.7\%, and linear probing from 68.0\% to 79.8\%.
Our method also achieves +4.8 mIoU ({\it i.e.,} 48.1 → 52.9) on ADE20K semantic segmentation, +6.6 $(\mathcal{J} \& \mathcal{F})_m$ ({\it i.e.,} 51.0 → 57.6) on DAVIS video segmentation, +2.5 AP$^{\mathtt{box}}$ ({\it i.e.,} 50.3 → 52.8) on COCO object detection, and +0.8 AP$^{\mathtt{mask}}$ ({\it i.e.,} 44.9 → 45.7) on COCO instance segmentation.
In addition, qualitative visualizations of collaborative attention vividly showcase the effectiveness of our CMT-MAE in learning meaningful representations. 
Extensive ablation studies also demonstrate the importance of collaborative masking and collaborative targets in learning masked autoencoders for improving downstream performance.

Our main contributions can be summarized as follows:

\begin{itemize}
   \item We present a simple yet effective masked autoencoder that can achieve collaborative masking and targets, called CMT-MAE, for boosting MIM-based visual pre-training.
   \item We propose a novel collaborative masking mechanism through linear aggregation across attention maps from both teacher and student networks to achieve powerful guidance.
   \item Extensive experiments comprehensively demonstrate the state-of-the-art superiority of our CMT-MAE over previous baselines on downstream tasks.
\end{itemize}

\section{Related Work}

\noindent\textbf{Self-supervised Visual Learning}.
Self-supervised visual learning aims to mine the internal characteristics from images without annotations by applying well-designed pretext tasks.
Early non-transformer researchers introduced instance-level~\cite{wu2018unsupervised,chen2020simple,chen2020big,grill2020bootstrap,he2019moco,chen2020mocov2,chen2021simsiam, jure2021barlow,wu2023accuracy,mo2023representation,mo2023exploring,wu2024rethinking} and cluster-based~\cite{caron2020unsupervised,li2021prototypical,wang2021cld,mo2021spcl,mo2022pauc} contrastive learning to pull representations from positive samples closer while pushing away features from negative pairs.
Recently, contrastive learning has been widely used in self-supervised vision transformers~\cite{chen2021mocov3,xie2021self-supervised,caron2021emerging,mo2023mcvt,mo2024dmtjepa,mo2024connecting} to achieve promising performance on visual downstream tasks. 
Typically, MoCov3~\cite{chen2021mocov3}  introduced a momentum encoder in ViT~\cite{dosovitskiy2021an} to minimize the distance between representations of two augmented views from the base encoder and momentum one.
To capture the local-to-global alignment, DINO~\cite{caron2021emerging} used a momentum encoder with multi-crop training to achieve knowledge distillation in the vision transformer.
In this work, our main focus is to learn meaningful visual representations in self-supervised transformers through another acclaimed technique, \ie, masked image modeling.

\begin{figure*}[t]
\centering
% \fbox{\rule{0pt}{4in}
% \rule{0.8\linewidth}{0pt}}
\includegraphics[width=0.96\linewidth]{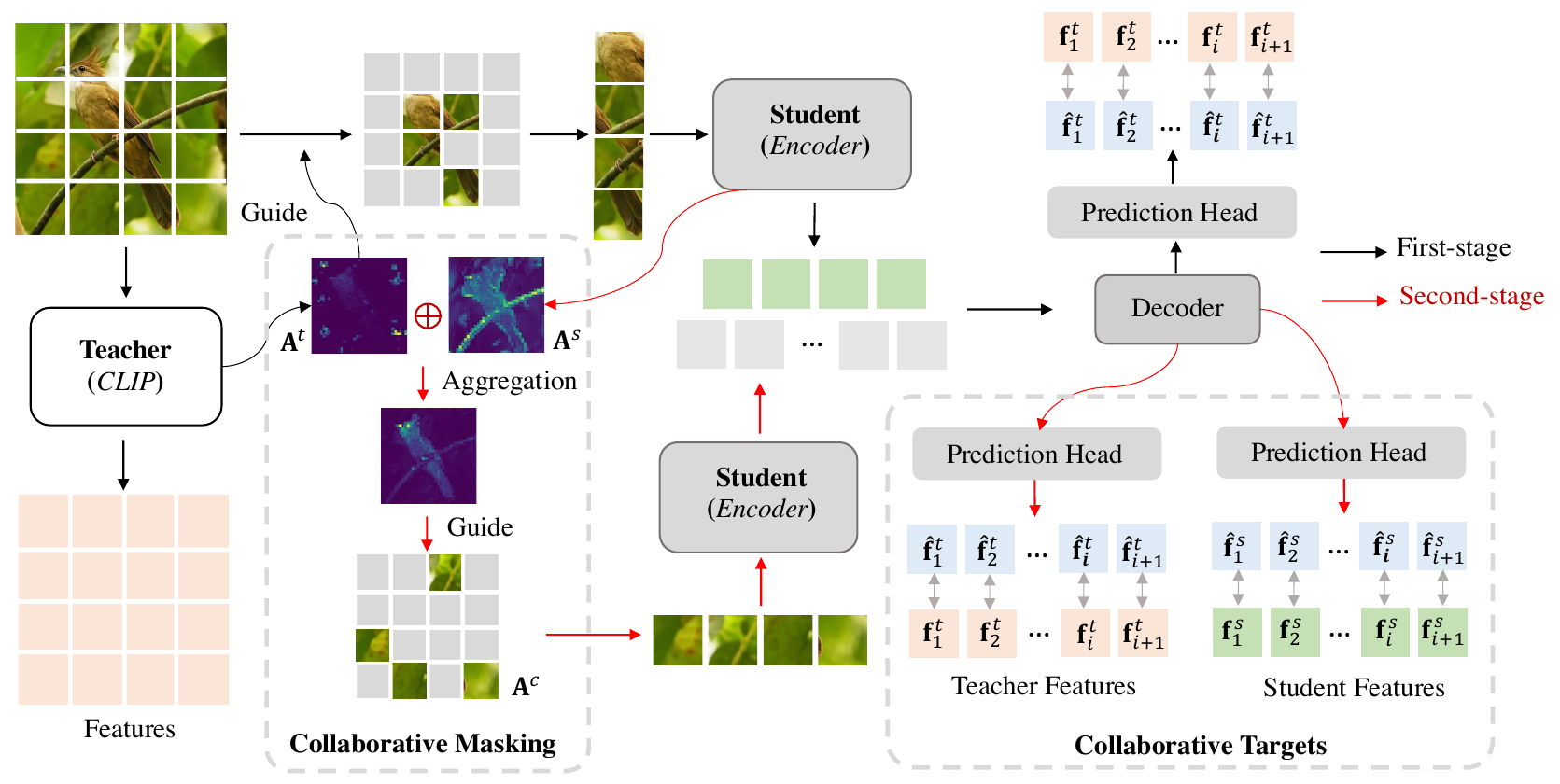}
\caption{Illustration of the proposed Masked Autoencoder with Collaborative Masking and Targets (CMT-MAE) framework.
{\bf First-stage:} a teacher transformer encoder (\ie, CLIP) takes an input image to extract an attention map $\bf{A}^t$ from the last attention layer to guide masking.
The student encoder generates features from unmasked patches, which are concatenated with masked tokens to feed into a decoder for recovering the teacher features $\mathbf{f}_i^t$ of masked patches.
{\red{\bf Second-stage:}}
a student momentum encoder takes the input image to generate a student-guided attention map $\bf{A}^s$, and linearly aggregates with a teacher-guided attention map $\bf{A}^t$ to produce the collaborative attention map $\bf{A}^c$ with a collaborative ratio $\alpha$ for collaborative masking.
Then masked tokens concatenate with features of unmasked patches from the student encoder to feed into the decoder.
Finally, two predicted heads are linearly applied to reconstruct the teacher features $\mathbf{f}_i^t$ and student features $\mathbf{f}_i^s$ of masked patches for collaborative targets.
Note that the collaborative ratio $\alpha$ is also applied to calculate collaborative losses from the teacher and student targets.
}
\label{fig: main_img}
\end{figure*}

\noindent\textbf{Masked Image Modeling}.
Masked image modeling (MIM) has been explored in many previous works~\cite{bao2021beit,atito2021sit,he2021masked,wei2022masked,xie2022SimMIM,wu2022objectwise,wu2023masked,wu2024dailymae} to reconstruct the masked image patch given the unmasked counterpart as clues. 
Early MIM approaches~\cite{bao2021beit,atito2021sit,he2021masked,li2021mst,shi2022adversarial} designed customized masking strategies (\eg, random, block-wise) as pre-text tasks during pre-training.
For example, block-wise masking was introduced in BEiT~\cite{bao2021beit} to learn transferrable visual representations by recovering discrete tokens of masked image patches.
Given features extracted from the 25\% unmasked patches, the seminal work, MAE~\cite{he2021masked} directly reconstructed missing pixels of 75\% masked patches.

Due to the benefit of open-sourced large-scale models, MVP~\cite{wei2022mvp} adopted CLIP~\cite{radford2021learning} pre-trained on 400 million image-language pairs to obtain multi-modal knowledge from images as guidance for supervision.
Based on iBoT~\cite{zhou2022ibot}, a distillation-based model, AttMask~\cite{kakogeorgiou2022attmask} extracted an attention map from the teacher transformer encoder to guide mask generation for pre-training, while SemMAE~\cite{li2022semmae} integrated semantic-guided masks from a self-supervised part-learning module with diversity constraints on attention.
However, they do not involve the self-training encoder itself (\ie, student) with dynamic knowledge from training data to help the teacher (\eg, CLIP) with fixed knowledge to guide mask and target generation for visual pre-training.
In contrast, we develop a novel collaborative masking mechanism to achieve dynamic and powerful guidance through a simple yet effective linear aggregation across attention maps from both teacher and student networks, which is not addressed before yet.

\section{Method}

Given an image with masked and unmasked patches, our target is to train a masked autoencoder framework with an encoder and a decoder to recover the masked patches using unmasked counterparts. 
We present a simple yet effective masked autoencoder with collaborative masking and targets, named CMT-MAE, which mainly consists of two modules, Collaborative Masking and Collaborative Targets.

\subsection{Preliminaries}

In this section, we first describe the problem setup and notations, and then revisit the masked image modeling in MAE~\cite{he2021masked} and teacher-guided MAE for self-supervised visual pre-training.

\noindent\textbf{Problem Setup and Notations.}
Given an image with a dimension of $3\times H\times W$ and a patch resolution of $P$, our goal is to learn a masked autoencoder framework with an encoder $f_e(\cdot)$ and a decoder $f_d(\cdot)$ to recover the masked patches using unmasked ones. 
We formally denote patch embeddings of raw input via each linear projection layer, \ie, $\mathbf{x}\in\mathbb{R}^{N\times D}$,
$H$ and $W$ are the height and width of each image, and $D$ is the dimension of features.
Note that $N = H/P\times W/P$ and $N$ is the total number of patches.

\noindent\textbf{Revisit Masked Autoencoder.}
To address the masked image modeling problem, MAE~\cite{he2021masked} first applied a random masking set $M$ along the total number of patches, and then an encoder to extract features from unmasked patches.
Finally, unmasked embeddings and masked tokens were concatenated into a decoder to recover the raw pixels of masked patches.
The vanilla masking loss for each image is calculated with the mean square loss between the targeted $\mathbf{p}_i$ and predicted normalized pixels $\hat{\mathbf{p}}_i$ as:  
\begin{equation}
    \mathcal{L}_{\text{mae}} = \dfrac{1}{|M|}\sum_{i\in M} ||\mathbf{p}_i - \hat{\mathbf{p}}_i||_2^2
\end{equation}
where $|M|$ denotes the total number of masked patches in the masking set $M$.

\noindent\textbf{Revisit Teacher-guided Masked Autoencoder.}
Based on the aforementioned seminal work, recent researchers~\cite{wei2022mvp,kakogeorgiou2022attmask,li2022semmae} started to use off-the-shelf large-scale pre-trained models (\eg, CLIP~\cite{radford2021learning}) as a teacher to guide mask generation and supervision targets. 
Specifically, they applied an attention map $\bf{A}^t$ extracted from a teacher transformer encoder to generate the masking set $M^t$, and representations as the reconstruction target.
The teacher-guided masking loss for each image is computed with the mean square loss between the targeted $\mathbf{f}_i^t$ and predicted normalized features $\hat{\mathbf{f}}_i^t$ as: 
\begin{equation}\label{eq:teacher-mae}
    \mathcal{L}_{\text{teacher-mae}} = \dfrac{1}{|M^t|}\sum_{i\in M^t} ||\mathbf{f}_i^t - \hat{\mathbf{f}}_i^t||_2^2
\end{equation}
where $|M^t|$ denotes the total number of masked patches in the teacher-guided masking set $M^t$.

However, such a training objective will pose the main challenge for collaborative masking and targets.
These approaches do not explicitly incorporate the self-training (\ie, student) model to cooperate with the teacher during pre-training.
Furthermore, the teacher and students have different knowledge levels: the teacher network (\eg, CLIP) has a higher-level knowledge than students at initial training; the student network starts to aggrandize its knowledge with more training epochs.
To address the challenge, we propose a simple yet effective masked autoencoder that can achieve collaborative masking and targets, called CMT-MAE, as illustrated in Figure~\ref{fig: main_img}.

\subsection{Collaborative Masking}\label{sec:cm}
In order to explicitly achieve collaborative masking guided by both the teacher and student, we propose a two-stage training paradigm with a simple collaborative masking mechanism through linear aggregation across attention maps from both models to improve the powerfulness of generated masks.
Specifically, in the first stage, we leverage a teacher transformer encoder (\ie, CLIP~\cite{radford2021learning}) takes an input image to extract an attention map $\bf{A}^t$ from the last attention layer to guide masking.
The student encoder generates features from unmasked patches, which are concatenated with masked tokens to feed into a decoder for recovering the teacher features $\mathbf{f}_i^t$ of masked patches.

Since momentum encoders~\cite{tarvainen2017mean,he2019moco} is a technique often used in self-supervised and semi-supervised learning to obtain slow-moving target representations, leading to more stable self-training and enhanced representations. 
In the second stage, as shown in Figure~\ref{fig: main_img}, we leverage a student momentum encoder to take the input image to generate a student-guided attention map $\bf{A}^s$.
Following~\cite{tarvainen2017mean,he2020momentum}, we update the student momentum encoder using an exponential moving average of the corresponding online encoders with coefficient $m$.\footnote{EMA update is $\hat{\theta} \leftarrow \hat{\theta} + (1-m) \theta$, where $\theta$ and $\hat{\theta}$ are the parameters of online and momentum encoders.}
The student $\bf{A}^s$ and teacher $\bf{A}^t$ attention maps are then linearly aggregated into a final collaborative map $\bf{A}^c$ of the form as
\begin{equation}
    \bf{A}^c = \alpha * \bf{A}^s + (1-\alpha) * \bf{A}^t
\end{equation}
where $\alpha$ denotes the collaborative ratio.
The collaborative map $\bf{A}^c$ is followingly applied to generate a masking set $M^c$ to split the input image into masked and unmasked patches for the second training stage.

\subsection{Collaborative Targets}\label{sec:ct}
Beyond collaborative masking, we introduce collaborative targets as the training objective in the second stage to dynamically select representations from both the teacher and student for simultaneous optimization.
With a masking set $M^c$ guided by the collaborative attention map $\bf{A}^c$, we concatenate masked tokens with features of unmasked patches from the student encoder to feed into the decoder.
Since updates to the student encoders are slowly incorporated into the momentum encoders, the target representations display smoother behavior during the training process.
In the end, we apply two predicted heads to linearly reconstruct the teacher features $\mathbf{f}_i^t$ and student features $\mathbf{f}_i^s$ of masked patches for collaborative targets.

\begin{table*}[t]
	%\normalem
	\renewcommand\tabcolsep{5.0pt}
	\centering
    \caption{{\bf ImageNet-1K image classification.} 
   We performed a linear probing and fine-tuning on pre-trained ViT-B/16 and ViT-L/16 models for image classification on ImageNet-1K benchmark.
   We report the top-1 accuracy to evaluate the quality of pre-trained representations. The best results are indicated in {\bf bold} numbers.}
   \label{tab: exp_sota_cls}
	\scalebox{0.6}{
		\begin{tabular}{lcccccc}
			\toprule
                \multirow{2}{*}{\bf Method} & \multirow{2}{*}{\bf Pre-train Dataset} & \multirow{2}{*}{\bf Pre-train Epochs } & \multicolumn{2}{c}{\bf ViT-B/16} & \multicolumn{2}{c}{\bf ViT-L/16} \\
			& & & \bf Linear Probing & \bf Fine-tuning & \bf Linear Probing & \bf Fine-tuning \\	
			\midrule
                \multicolumn{7}{l}{\it Training from scratch} \\
                 DeiT~\cite{touvron2020deit}  & -- & -- & -- & 81.8 & -- & -- \\
                 ViT~\cite{dosovitskiy2021an}  & -- & -- & -- & 82.3 & -- & 84.5 \\ \hline
                \multicolumn{7}{l}{\it Contrastive-based Pre-Training} \\
                AttMask~\cite{kakogeorgiou2022attmask} & ImageNet-1K & 100 & 75.7 & -- & -- & -- \\
                DINO~\cite{caron2021emerging} & ImageNet-1K & 300 & 78.2 & 82.8 & -- & -- \\
                MoCo v3~\cite{chen2021mocov3} & ImageNet-1K & 300 & 76.5 & 83.2 & -- & 84.1 \\
                iBOT~\cite{zhou2022ibot} & ImageNet-1K & 1600 & 79.5 & 84.0  & 81.0 & 84.8 \\ \hline
                \multicolumn{7}{l}{\it MIM-based Pre-Training} \\
                BEiT~\cite{bao2021beit} & ImageNet-1K & 800 & 56.7 & 83.2 & -- & -- \\
                MAE~\cite{he2021masked} & ImageNet-1K & 1600 & 68.0 & 83.6 & 75.1 & 85.9 \\
                MaskFeat~\cite{wei2022masked} & ImageNet-1K & 1600 & 68.0 & 84.0 & -- & 85.7 \\
                SimMIM~\cite{xie2022SimMIM} & ImageNet-1K & 800 & 56.7 & 83.8 & -- & -- \\
                PeCo~\cite{dong2021peco} & ImageNet-1K & 300 & -- & 84.1 & -- & -- \\
                MVP~\cite{wei2022mvp} & ImageNet-1K & 300 & -- & 84.4 & -- & 86.3 \\
                data2vec~\cite{alexei2022data2vec} & ImageNet-1K & 1600 & -- & 84.2 & -- & 86.6 \\
                SemMAE~\cite{li2022semmae} & ImageNet-1K & 800 & 68.7 & 84.5 & -- & -- \\
                \rowcolor{blue!10}
                CMT-MAE (ours) & ImageNet-1K & 800 & \textbf{79.8} & \textbf{85.7} &  \textbf{81.6} & \textbf{87.2} \\
			\bottomrule
			\end{tabular}}
\end{table*}

The overall objective of our model in the second stage is simply optimized in an end-to-end manner as:
\begin{equation}
    \mathcal{L}_{\text{cmt-mae}} = \dfrac{1}{|M^c|}\sum_{i\in M^c} \alpha * ||\mathbf{f}_i^s - \hat{\mathbf{f}}_i^s||_2^2 + (1-\alpha) * ||\mathbf{f}_i^t - \hat{\mathbf{f}}_i^t||_2^2
\end{equation}
where $|M^c|$ denotes the total number of masked patches in the collaborative masking set $M^c$.
$\mathbf{f}_i^s, \hat{\mathbf{f}}_i^s$ denote the target and prediction of student features, and $\mathbf{f}_i^t, \hat{\mathbf{f}}_i^t$ for teacher features.
$\alpha$ is the collaborative ratio.
Note that the collaborative ratio $\alpha$ is applied to calculate collaborative losses from the teacher and student targets.

\section{Experiments}

\subsection{Experimental setup}

\noindent\textbf{Datasets.}
Following previous methods~\cite{he2021masked}, we use ImageNet-1K~\cite{imagenet_cvpr09} for image classification, MS-COCO~\cite{lin2014coco} for object detection and instance segmentation, and ADE20K~\cite{zhou2017scene,Zhou2018SemanticUO} for semantic segmentation.
We closely follow previous work~\cite{chen2021mocov3,xie2021self-supervised,caron2021emerging}, and adopt the Mask R-CNN~\cite{he2017mask} as the detector. 
The ViT-Base~\cite{dosovitskiy2021an} backbone weights are initialized with weights pre-trained on ImageNet-1K using our CMT-MAE. 
Other settings are the same as the implementation in this work~\cite{he2021masked}.
% \noindent \textbf{ADE20K.}
Following the settings in~\cite{he2021masked,bao2021beit}, we use the UPerNet approach~\cite{xiao2018unified} based on our ImageNet-1K pre-trained ViT-Base for evaluation.
For a fair comparison, we fine-tune the detector with the same learning rate in~\cite{he2021masked,bao2021beit}.
For video object segmentation, we use DAVIS-2017 dataset~\cite{ponttuset2017davis} that includes 60 training, 30 validation, and 60 testing videos.

\begin{table}[t]
	%\normalem
	\renewcommand\tabcolsep{12.0pt}
	\centering
        \caption{{\bf COCO object detection, instance segmentation, and ADE20K semantic segmentation.} We fine-tuned pre-trained ViT-B/16 models to perform COCO object detection, instance segmentation, and ADE20K semantic segmentation. 
   The AP$^{\mathtt{box}}$, AP$^{\mathtt{mask}}$, and mIoU metrics denote the results of COCO detection, COCO segmentation, and ADE20K segmentation, respectively.
    The best results are indicated in {\bf bold} numbers.}
   \label{tab: exp_sota_det_seg}
	\scalebox{0.75}{
		\begin{tabular}{lccc}
			\toprule
                \bf Method & \bf AP$^{\mathtt{box}}$ & \bf AP$^{\mathtt{mask}}$ & \bf mIoU \\	
			\midrule
                \multicolumn{4}{l}{\it Supervised Training}
                 \\
                 DeiT~\cite{touvron2020deit} & 47.9 & 42.9 & 47.4 \\ \hline
                 \multicolumn{4}{l}{\it Contrastive-based Pre-Training}
                 \\
                 DINO~\cite{caron2021emerging} & 46.8 & 41.5 & 47.2 \\
                 MoCo v3~\cite{chen2021mocov3} & 47.9 & 42.7 & 47.3 \\ \hline
                 \multicolumn{4}{l}{\it MIM-based Pre-Training}
                 \\
                 BEiT~\cite{bao2021beit} & 42.1 & 37.8 & 45.8 \\
                 MAE~\cite{he2021masked} (800epoch) & 48.4 & 42.6 & 46.1 \\ 
                 
                 PeCo~\cite{dong2021peco} & 43.9 & 39.8 & 46.7 \\
                 SplitMask~\cite{alaaeldin2021splitmask} & 46.8 & 42.1 & 45.7\\
                 iBoT~\cite{zhou2022ibot} & 48.2 & 42.7 & 50.0 \\
                 CAE~\cite{chen2022cae} & 49.2 & 43.3 & 48.8 \\
                 MVP~\cite{wei2022mvp} & -- & -- & 52.4 \\
                 SemMAE~\cite{li2022semmae} & -- & -- & 46.3 \\
                 MAE~\cite{he2021masked} (1600epoch) & 50.3 & 44.9 & 48.1 \\
                 \rowcolor{blue!10}
                 CMT-MAE (ours) & \bf 52.8 & \bf 45.7 & \bf 52.9 \\
			\bottomrule
			\end{tabular}}
\end{table}

\noindent\textbf{Evaluation Metrics.}
We follow previous masked image modeling work~\cite{he2021masked,bao2021beit} to report the classification accuracy of linear probing and fine-tuning. 
For object detection and instance segmentation on MS-COCO, we apply AP$^b$ and AP$^m$ as metrics for the bounding boxes and the instance masks.
mIoU results are reported to evaluate semantic segmentation on ADE20K. 
For video object segmentation on DAVIS-2017, we use Jabri-based $(\mathcal{J} \& \mathcal{F})_m$, $\mathcal{J}_m$, $\mathcal{F}_m$ as metrics to evaluate segmenting frames based on the nearest neighbor between consecutive scenes.

\begin{figure*}[t]
\centering
% \fbox{\rule{0pt}{2in}
% \rule{0.8\linewidth}{0pt}}
\includegraphics[width=0.75\linewidth]{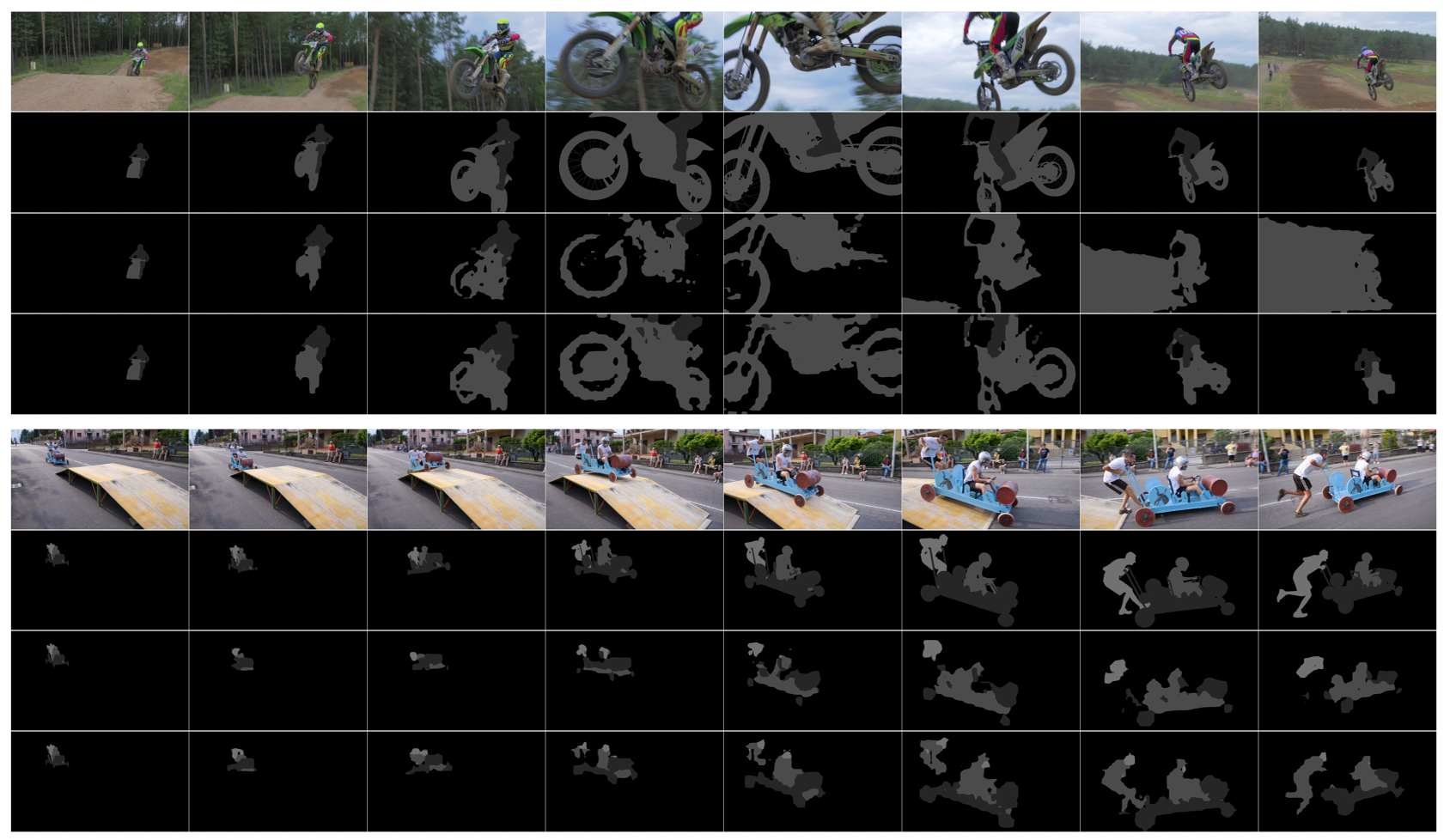}
\caption{{\bf Visualizations of DAVIS 2017 video object segmentation.}
Four rows for each case represent raw frames, ground-truth masks, MAE predictions, and our CMT-MAE predictions.
We visualize the segmentation masks of DAVIS 2017 video object segmentation using ViT-B/16 pre-trained on ImageNet-1K. 
The proposed CMT-MAE produces much more accurate and high-quality segmentation masks.}
\label{fig: vis_seg_video}
\end{figure*}

\noindent\textbf{Implementation.}
For input images, the resolution is resized to $224 \times 224$, \ie, $H=W=224$. 
We follow prior work~\cite{he2021masked} and apply a patch size of $16$, \ie, $P=16$.
The base and large models of ViT~\cite{dosovitskiy2021an} architecture are used for experiments.
We mask 75\% on patches of each image, same as in MAE~\cite{he2021masked}.
The model is pre-trained on ImageNet-1K for 800 epochs with the AdamW~\cite{loshchilov2018decoupled} optimizer with a learning rate of 1.5e-4, a decay rate of 0.05, and a batch size of 4096.
For fine-tuning on ImageNet-1K, the model is trained for 100 epochs with a batch size of 256.
For the MS-COCO dataset, we train the model for 12 epochs with a batch size of 16, and an initial learning rate of 2e-4. 
The learning rate is decayed by 10 at epochs 8 and 11.
For the ADE20K dataset, the model is trained for 160K iterations with an initial learning rate of 3e-5 and a layer-wise learning rate decay of 0.9. 
We set the weight decay to 0.05 and the drop path rate to 0.1.
Multi-scale testing is not used for a fair comparison with precious approaches~\cite{he2021masked,bao2021beit}.

\begin{table}[t]
	%\normalem
	\renewcommand\tabcolsep{6.0pt}
	\centering
    \caption{\textbf{DAVIS video object segmentation.} 
   We performed DAVIS 2017 video object segmentation using ViT-B/16 and ViT-L/16 pre-trained on ImageNet-1K. 
         We report Jabri-based metrics $(\mathcal{J} \& \mathcal{F})_m$, $\mathcal{J}_m$, $\mathcal{F}_m$ to evaluate the quality of frozen pre-trained representations.
         The best results are indicated in {\bf bold} numbers.
         }
    \label{tab: exp_sota_seg_video}
	\scalebox{0.75}{
		\begin{tabular}{lccccc}
			\toprule
			\bf Method & \bf Backbone & $\boldsymbol{(\mathcal{J} \& \mathcal{F})_m}$ & $\boldsymbol{\mathcal{J}_m}$ & \bf $\boldsymbol{\mathcal{F}_m}$  \\
   \midrule
   % \multicolumn{4}{l}{\it MIM-based Pre-Training} \\
    MAE~\cite{he2021masked} & ViT-B/16 & 51.0 & 49.4 & 52.6 \\
    I-JEPA~\cite{assran2023self} & ViT-B/16 & 56.2 & 56.1 & 56.3 \\
    \rowcolor{blue!10}
    CMT-MAE (ours) & ViT-B/16 & \bf 57.6 & \bf 56.7 & \bf 58.5 \\ \hline
    MAE~\cite{he2021masked} & ViT-L/16 & 53.4 & 52.5 & 54.3 \\
    I-JEPA~\cite{assran2023self} & ViT-L/16 & 56.6 & 56.3 & 56.9 \\
    \rowcolor{blue!10}
   CMT-MAE (ours) & ViT-L/16 & \bf 60.5 & \bf 59.7 & \bf 61.3 \\
   \bottomrule
			\end{tabular}}
\end{table}

\begin{table*}[t]
	%\normalem
	\renewcommand\tabcolsep{12.0pt}
	\centering
           \caption{{\bf Ablation studies on component analysis.} We performed ablation studies on Collaborative Masking (CM) and Collaborative Targets (CT) modules using a pre-trained ViT-B/16 on ImageNet-1K. The best results are indicated in {\bf bold} numbers.}
   \label{tab: ab_module}
	\scalebox{0.6}{
		\begin{tabular}{cccccccccc}
			\toprule
		\bf CM & \bf CT & \bf Linear Probing & \bf Fine-tuning &  \bf AP$^{\mathtt{box}}$ & \bf AP$^{\mathtt{mask}}$ & \bf mIoU & $\boldsymbol{(\mathcal{J} \& \mathcal{F})_m}$ & $\boldsymbol{\mathcal{J}_m}$ & \bf $\boldsymbol{\mathcal{F}_m}$  \\
			\midrule
			\xmark & \xmark & 68.0 & 83.6 & 48.4 & 42.6 & 46.1 & 51.0 & 49.4 & 52.6 \\
               \cmark & \xmark & 73.5 & 84.2 & 50.3 & 43.5 & 48.3 & 53.8 & 51.8 & 55.8 \\
             \xmark & \cmark & 74.2 & 84.5 & 50.9 & 44.2 & 49.2 & 54.6 & 52.9 & 56.3 \\
               \rowcolor{blue!10}
               \cmark & \cmark & \bf 79.8 & \bf 85.7 & \bf 52.8 & \bf 45.7 & \bf 52.9 & \bf 57.6 & \bf 56.7 & \bf 58.5 \\
			\bottomrule
			\end{tabular}}
\end{table*}

\begin{table*}[t]
	%\normalem
	\renewcommand\tabcolsep{12.0pt}
	\centering
        \caption{{\bf Ablation studies on collaborative ratio $\alpha$.} We performed ablation studies using an ImageNet-1K pre-trained ViT-B/16 model to explore impacts on collaborative ratio. The best results are indicated in {\bf bold} numbers. }
   \label{tab: ab_ratio}
	\scalebox{0.65}{
		\begin{tabular}{ccccccccc}
			\toprule
		\bf $\alpha$ & \bf Linear Probing & \bf Fine-tuning &  \bf AP$^{\mathtt{box}}$ & \bf AP$^{\mathtt{mask}}$ & \bf mIoU & $\boldsymbol{(\mathcal{J} \& \mathcal{F})_m}$ & $\boldsymbol{\mathcal{J}_m}$ & \bf $\boldsymbol{\mathcal{F}_m}$  \\
			\midrule
                0\% & 77.1 & 84.6 & 51.7 & 44.7 & 51.5 & 55.3 & 54.0 & 56.6 \\
			10\% & 78.7 & 85.1 & 52.1 & 45.2 & 52.1 & 56.5 & 55.6 & 57.4 \\
                \rowcolor{blue!10}
                30\% & \bf 79.8 & \bf 85.7 & \bf 52.8 & \bf 45.7 & \bf 52.9 & \bf 57.6 & \bf 56.7 & \bf 58.5 \\
                50\% & 79.3 & 85.5 & 52.5 & 45.5 & 52.5 & 57.2 & 56.1 & 58.3 \\
                70\% & 78.9 & 85.2 & 52.2 & 45.3 & 52.3 & 56.8 & 55.8 & 57.8 \\
                90\% & 78.3 & 84.9 & 51.9 & 45.1 & 51.9 & 56.2 & 55.3 & 57.1 \\
                100\% & 77.6 & 84.7 & 51.8 & 44.9 & 51.7 & 55.8 & 54.7 & 56.9 \\
			\bottomrule
			\end{tabular}}
\end{table*}

\subsection{Comparison to prior work}

In this work, we propose a novel and effective framework with MAE pre-training for downstream tasks, \ie, linear probing, fine-tuning, object detection, instance segmentation, semantic segmentation, and video object segmentation. 
To validate the effectiveness of the proposed CMT-MAE, we comprehensively compare it to previous baselines, including training from scratch~\cite{touvron2020deit,dosovitskiy2021an}, contrastive-based pre-training~\cite{kakogeorgiou2022attmask,caron2021emerging,chen2021mocov3,zhou2022ibot}, and previous MIM-based pre-training~\cite{bao2021beit,he2021masked,wei2022masked,xie2022SimMIM,dong2021peco,wei2022mvp,alexei2022data2vec,li2022semmae} approaches.

\noindent {\bf Image classification.} 
Table~\ref{tab: exp_sota_cls} reports the quantitative comparison results of linear probing and fine-tuning on pre-trained ViT-B/16 and ViT-L/16 models.
As can be seen, we achieve the best performance in terms of all metrics for both models.
In particular, the proposed CMT-MAE significantly outperforms MAE~\cite{he2021masked}, the original masked image modeling baseline, by 11.8\% \& 2.1\% and 6.5\% \& 1.3\% relative top-1 accuracies in terms of linear probing \& fine-tuning on ViT-B/16 and ViT-L/16 models.
Moreover, we achieve superior performance gains compared to SemMAE~\cite{li2022semmae}, the recent MIM-based pre-training approach that applied semantic-guided masks and diversity constraints on attention.
Meanwhile, our CMT-MAE outperforms iBoT~\cite{zhou2022ibot} by 1.7\% and 2.4\% relative top-1 accuracies in terms of fine-tuning on ViT-B/16 and ViT-L/16 models.
The proposed CMT-MAE also achieves better results than DINO~\cite{caron2021emerging}, a strong contrastive-based pre-training baseline. 
These significant improvements demonstrate the superiority of our method in learning better representations during pre-training for image classification.

\noindent {\bf Object detection \& instance segmentation.} 
For the COCO object detection \& instance segmentation benchmarks, we report the quantitative comparison results of COCO object detection and instance segmentation on the pre-trained ViT-B/16 model in Table~\ref{tab: exp_sota_det_seg}. 
We can observe that the proposed CMT-MAE achieves the best results in terms of all metrics.
Compared to MAE~\cite{he2021masked} trained on 1600 epochs, we achieve performance gains of 2.5@AP$^{\mathtt{box}}$ and 0.8@AP$^{\mathtt{mask}}$.
We also achieve highly better results than other contrastive-based~\cite{caron2021emerging} and MIM-based~\cite{bao2021beit,li2022semmae} pre-training approaches.

\noindent {\bf Semantic segmentation.} 
Table~\ref{tab: exp_sota_det_seg} also shows the quantitative comparison results of ADE20K semantic segmentation on the ViT-B/16 model pre-trained on ImageNet-1K.
Our CMT-MAE significantly outperforms MAE~\cite{he2021masked} by 4.8@mIoU and also achieves better performance than MVP~\cite{wei2022mvp}, the strong baseline using CLIP~\cite{radford2021learning} knowledge pre-trained on 400 million image-text pairs.
These results further validate the effectiveness of our collaborative masking and collaborative masking in learning discriminative embeddings across pre-training for semantic segmentation.

\noindent {\bf Video object segmentation.} 
We also present additional video object segmentation on the DAVIS 2017 benchmark using ImageNet-1K pre-trained ViT-B/16 and ViT-L/16 models, as shown in Table~\ref{tab: exp_sota_seg_video}.
We achieve superior performance gains of 6.6@$(\mathcal{J} \& \mathcal{F})_m$,  7.3@$\mathcal{J}_m$, 5.9@$\mathcal{F}_m$, and  7.1@$(\mathcal{J} \& \mathcal{F})_m$, 7.2@$\mathcal{J}_m$, 7.0@$\mathcal{F}_m$ in terms of pre-trained ViT-B/16 and ViT-L/16, compared to MAE~\cite{he2021masked}.
We also achieve better results than I-JEPA~\cite{assran2023self}, the recent joint embedding predictive architecture for image self-supervised learning.
These qualitative results also showcase the effectiveness of our CMT-MAE in generating much more accurate and high-quality segmentation masks on video objects compared to MAE~\cite{he2021masked}, as shown in Figure~\ref{fig: vis_seg_video}.

\subsection{Experimental analysis}

In this section, we performed ablation studies to demonstrate the benefit of introducing Collaborative Masking (CM) and Collaborative Targets (CT) modules. 
We also conducted extensive experiments to explore the impact of collaboration ratio $\alpha$ and learned meaningful collaborative attention maps.

\noindent\textbf{Collaborative Masking \& Collaborative Targets.}
In order to validate the effectiveness of the introduced collaborative masking (CM) and collaborative targets (CT), we ablate the necessity of each module and report the quantitative results in Table~\ref{tab: ab_module}.
We can observe that adding CM to the vanilla baseline highly increases the results of 5.5@Linear Probing, 0.6@Fine-tuning, 1.9@AP$^{\mathtt{box}}$, 0.9@AP$^{\mathtt{mask}}$, 2.2@mIoU, 2.8@$(\mathcal{J} \& \mathcal{F})_m$,  2.4@$\mathcal{J}_m$, 3.2@$\mathcal{F}_m$, which demonstrates the benefit of collaborative masking in generating meaningful representations guided by both the teacher and student during pre-training.
Meanwhile, introducing only CT in the baseline also increases the downstream performance in terms of all metrics.
More importantly, incorporating CM and CT together into the baseline significantly raises the performance by 11.8 @Linear Probing, 2.1@Fine-tuning, 4.4@AP$^{\mathtt{box}}$, 3.1@AP$^{\mathtt{mask}}$, 6.8@mIoU, 6.6@$(\mathcal{J} \& \mathcal{F})_m$,  7.3@$\mathcal{J}_m$, 5.9@$\mathcal{F}_m$. 
These improving results validate the importance of collaborative masking and targets in learning collaborative representations from both teacher and student for masked autoencoders.

\noindent\textbf{Trade-off on Collaborative Ratio.}
The number of collaborative ratios in the proposed collaborative masking and targets affects the pre-trained representations for diverse downstream tasks.
To explore such effects more comprehensively, we varied the number of ratios from $\{0\%,10\%,30\%,50\%,70\%,90\%,100\%\}$.
The comparison results of all downstream tasks using a ViT-B/16 model pre-trained on ImageNet-1K are reported in Table~\ref{tab: ab_ratio}.
When the number of collaboration ratio $\alpha$ is $30\%$, we achieve the best downstream performance in terms of all metrics. 
With the increase of
collaboration ratio from $0\%$ to $30\%$, the proposed CMT-MAE consistently raises results, which shows the importance of collaborative masking and collaborative targets in masked autoencoders for learning discriminative representations. 
However, increasing the collaboration ratio from $30\%$ to $90\%$ will not continually improve the result since there might be a trade-off between the teacher and student to learn different representations during pre-training.

\section{Conclusion}

In this work, we present CMT-MAE, a simple yet effective masked autoencoder that can simultaneously achieve collaborative masking and targets.
We leverage a novel collaborative masking mechanism through linear aggregation across attentions from both teacher and student models.
We further use their representations as the collaborative target of the decoder for reconstruction.
Experimental results on ImageNet-1K, MS-COCO, ADE-20K, and DAVIS 2017 validate the state-of-the-art superiority of the proposed framework. 
In addition, qualitative visualizations vividly showcase the effectiveness of our CMT-MAE in capturing meaningful representations for downstream tasks. 
Extensive ablation studies also demonstrate the importance of collaborative masking and collaborative targets in learning masked autoencoders for improving downstream performance.

\bibliography{reference}

\end{document}